\documentclass[11pt]{article}

\usepackage[final]{acl}
\usepackage{times}
\usepackage{latexsym}
\usepackage[T1]{fontenc}
\usepackage[utf8]{inputenc}
\usepackage{microtype}
\usepackage{inconsolata}
\usepackage{booktabs}
\usepackage{tabularx}
\usepackage{array}
\usepackage{graphicx}
\usepackage{multirow}
\usepackage{amsmath}
\usepackage{tikz}
\usetikzlibrary{arrows.meta,positioning,calc}

\definecolor{piNavy}{HTML}{17324D}
\definecolor{piTeal}{HTML}{2B7A78}
\definecolor{piRule}{HTML}{9AAAB8}
\definecolor{piRuntime}{HTML}{E3F3F1}
\definecolor{piModel}{HTML}{E7F0E8}
\definecolor{piData}{HTML}{EDF2F6}
\definecolor{piModelGreen}{HTML}{477A57}

\newcommand{\SearchAgent}{\textsc{Search}}
\newcommand{\ResearchAgent}{\textsc{Research}}

\tikzset{
  piBox/.style={
    draw=piRule,
    fill=piData,
    rounded corners=3pt,
    align=center,
    text=piNavy,
    font=\small,
    inner sep=6pt
  },
  piRuntimeBox/.style={piBox,fill=piRuntime,draw=piTeal},
  piModelBox/.style={piBox,fill=piModel,draw=piModelGreen},
  piDataBox/.style={piBox,fill=piData,draw=piRule},
  piFlow/.style={-{Stealth[length=2mm]},draw=piTeal,thick}
}

\title{TRU-PULSE: \underline{P}opulation \underline{U}nderstanding through \underline{L}ongitudinal \underline{S}ynthesis of \underline{E}vidence}

\author{
  Mohammadmahdi Shokri \\
  Truveta
  \And
  Bhargav Shandilya \\
  Truveta
  \And
  Mariyam Amir \\
  Truveta
  \AND
  Ashkan Jeddi \\
  Truveta
  \And
  Saman Zarandioon \\
  Truveta
}

\begin{document}
\maketitle

\begin{abstract}
Real-world clinical data is longitudinal. Over many visits, patients build
long histories of notes, diagnoses, and treatments that together form a
patient journey. These journeys enable investigation of longitudinal clinical
questions that link earlier conditions, exposures, or interventions to later
outcomes. Answering such questions is difficult because relevant evidence is
distributed across temporally distant phases of a patient's journey, and no
single note contains it.

We present TRU-PULSE, an agentic system for extracting insights
grounded in real clinical journeys. TRU-PULSE divides the task into
\SearchAgent{} and \ResearchAgent{}. \SearchAgent{} maps a natural-language
question to clinical concepts, finds relevant parts of each patient's journey,
and combines them into patient-level evidence. \ResearchAgent{} then reasons
over this evidence, synthesizes findings with note-level citations, and
assesses whether the evidence is sufficient. We evaluate TRU-PULSE on a
large proprietary real-world clinical records and a synthetic dataset.
Our results show hybrid ranking over patient-level summaries is the most effective search
strategy. \ResearchAgent{} produces findings grounded in cited source notes
and flags questions with insufficient evidence. Clinical experts verify 1.00
patient precision and 0.97 note precision. On 17 questions run over real
clinical data, all answer-relevance and clinical-accuracy ratings are 5/5.
\end{abstract}

\section{Introduction}
\label{sec:intro}
 
Clinical notes contain comprehensive information about patient care:
how a disease progressed, why a clinician changed treatment, and whether an intervention produced the intended response.  
Over months or years of care, these notes accumulate into a \emph{patient journey}, a longitudinal record that 
enables research on disease progression and treatment response, building cohorts of patients with similar trajectories, 
and proactively surfacing patterns that motivate new studies. 
Realizing this potential depends on access to large, comprehensive, diverse, and 
longitudinal datasets of clinical notes spanning multiple patient visits. 
However, these datasets are difficult to obtain due to strict privacy governance, 
data-sharing regulations, and fragmentation across disparate health systems. 
Scale poses another significant challenge; even when large-scale data is available, 
it remains nearly impossible for researchers or clinicians to manually identify the 
correct patient cohorts for a given clinical query, 
and sending all the records through large language models (LLM) is computationally impractical. 
Furthermore, while these detailed notes are invaluable for clinical research, their unstructured nature, 
characterized by excessive length while often containing repeated or redundant information, 
makes extracting relevant insights highly challenging.
Finally, Longitudinal research relies on queries that are fundamentally more complex than conventional 
retrieval. A query such as \emph{``find patients with diabetes''} can be answered by matching a 
single note. By contrast, \emph{``which patients developed renal complications after starting a 
new therapy, and what preceded them?''} requires connecting evidence from temporally distant parts of a 
journey that may span many encounters, specialties, and note types. A patient is relevant only if 
the combination of their notes supports the query, and surface matches may be negated, historical, 
uncertain, or describe plans that were never carried out

These requirements motivate a division between patient discovery and evidence interpretation. 
As such, we introduce TRU-PULSE,\footnote{The source code and Git repository will be released publicly upon acceptance.} an agentic 
\textsc{Search}--\textsc{Research} system for extracting evidence-grounded insights from large collections of longitudinal clinical records.

In the discovery stage, the \textsc{Search} agent asks \emph{where should the system look?} 
This is a decisive stage: the agent must efficiently search across 
large patient populations to identify the cohort relevant to a given query, as a relevant patient missed during 
search cannot be recovered downstream. To enable this, the \textsc{Search} agent translates the natural-language 
query into clinically meaningful concepts and searches compact longitudinal patient representations, 
allowing evidence distributed across multiple encounters to jointly surface relevant patients without 
placing every patient history into the model's context.

The \textsc{Research} agent then asks \emph{what do the selected patients' records support?} 
This is the evidence interpretation stage, where more computation can be devoted to the much smaller candidate cohort. 
Building on the language understanding and reasoning capabilities of LLMs, the agent returns to the original note text, 
identifies relevant evidence, and synthesizes an answer with note-level citations and an explicit sufficient, partial, 
or insufficient evidence status. Together, the \textsc{Search} and \textsc{Research} 
stages provide an end-to-end system for generating evidence-grounded insights from longitudinal clinical records while maintaining 
a traceable connection between each finding and its source.

We evaluate TRU-PULSE on a large real-world corpus of approximately 2.1M clinical notes from 33K patients, 
and a controlled synthetic dataset. Our experiments show that retrieval representation is a central design decision for patient discovery:
hybrid ranking over note summaries consistenly yields the most precise candidate cohorts across settings.
Building on these cohorts, the \textsc{Research} agent grounds its finding in cited source notes and distinguishes questions
the records fully support from those with only partial or insufficient evidence.

We make four key contributions. First, we introduce a patient-first SEARCH agent that discovers 
longitudinally relevant patient cohorts across millions of clinical notes, identifying individuals 
whose relevance depends on evidence scattered throughout their medical history. Second, we 
introduce a source-grounded RESEARCH agent that evaluates selected cohorts using original 
clinical notes, complete with note-level citations and explicit evidence-sufficiency states. 
Third, we present TRU-PULSE, an end-to-end agentic system that links both agents via an 
evidence-bounded workflow to enable scalable, auditable longitudinal analysis from a single 
natural-language query. Finally, we perform a rigorous testing using real and synthetic health-care data, 
and evaluation using both large language models (LLMs) and clinical experts to validate system performance. 
While the search and research components are integrated here for unified insight generation, 
each agent offers distinct clinical advantages: the Search agent can be used independently 
for cohort selection, study design, and model building, while the Research agent can 
be deployed on its own to extract deep, focused insights from existing patient journeys.

\section{Related Work} 
\label{sec:related-work} 
Prior clinical benchmarks capture complementary aspects of patient retrieval and question answering. 
The TREC Medical Records Track evaluates cohort discovery from clinical records \citep{voorhees2012trec}, 
while CliCR and emrQA study question answering over clinical case reports and clinical notes, 
respectively \citep{suster2018clicr,pampari2018emrqa}. 
These benchmarks motivate reasoning over clinical text, 
but do not jointly address patient-level retrieval and evidence-grounded reasoning across longitudinal records. 

Related work has also explored computable representations of longitudinal electronic health records (EHRs). 
Kim et al. \citep{kim2026computable} integrate structured and unstructured EHR data into provenance-linked, 
ontology-grounded knowledge graphs for longitudinal analysis. 
Our approach is complementary: rather than transforming the corpus into a persistent structured representation, 
we retrieve patients through compact longitudinal representations and return to their source notes for 
evidence-grounded reasoning. 

More broadly, retrieval-augmented generation (RAG) conditions language models on retrieved external 
evidence \citep{lewis2020retrieval}, with retrieval commonly operating over passages 
using sparse or dense methods such as BM25 and DPR \citep{robertson2009bm25,karpukhin2020dense}. 
Longitudinal clinical reasoning presents a different retrieval unit: relevant evidence may span multiple 
encounters, note types, and dates for the same patient. We therefore frame retrieval at the patient level while 
grounding subsequent reasoning in the underlying notes. 

Finally, work on attributed generation has emphasized 
linking generated claims to supporting evidence, including post-hoc attribution and revision in RARR \citep{gao2023rarr} 
and citation-aware generation and evaluation in ALCE \citep{gao2023alce}. 
TRU-PULSE shares this goal while constraining evidence to the resolved patient scope and retaining 
note-level provenance and explicit evidence-sufficiency states. 
This makes patient discovery, evidence interpretation, and answer synthesis separately observable and evaluable.

\section{Task and System Overview}
\label{sec:overview}

\begin{figure}[t]
\centering
\includegraphics[width=0.8\columnwidth]{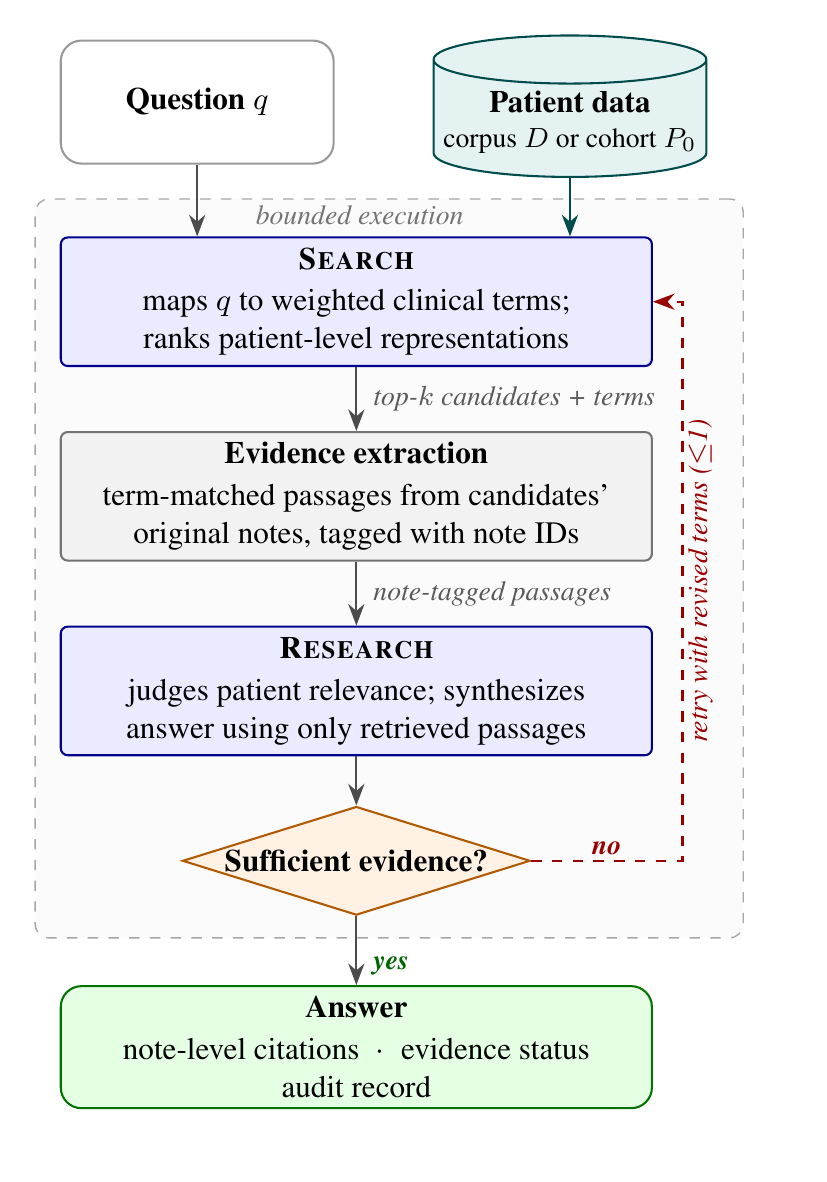}
\caption{TRU-PULSE workflow.}
\label{fig:workflow}
\end{figure}
 
\paragraph{Task definition.}
TRU-PULSE answers a natural-language clinical question over a
specified longitudinal-record corpus. A request contains a corpus
$\mathcal{D}$ and a clinical question $q$. The system identifies candidate patients, evaluates their
source notes within that scope, and returns an answer, note-level
citations, an evidence status, and audit metadata. The questions we
target are longitudinal: answering them requires combining evidence from
different parts of a patient's journey, so a retrieval match on any single
note does not establish the requested clinical relationship.
 
\paragraph{Architecture.}
TRU-PULSE divides each request between two agents.
The \textsc{Search} agent ranks compact patient-level representations of
the corpus, yielding a bounded candidate set. The \textsc{Research} agent
then returns to the selected patients' original notes, identifies relevant
evidence, and synthesizes an answer with note-level citations. A
deterministic runtime enforces patient scope, budgets, and citation
resolution. When the retrieved records do not support an answer, the
\textsc{Search} agent may re-plan once with revised terms; otherwise the
system reports insufficient evidence rather than producing an
unsupported answer.
 
The \textsc{Search} agent follows a plan-then-execute design. In the
planning step, a language model interprets $q$, assigns a query profile,
and emits scored core and related clinical terms. This plan serves as a
typed call to the search tool, with the query profile setting the
retrieval budget $k$. In the execution step, the agent invokes the search
tool over a patient-level index built offline from the corpus. The tool
supports several ranking algorithms, and we use the hybrid algorithm by
default based on our comparison in Section~\ref{sec:ablations}.

\section{\SearchAgent{} Agent}
\label{sec:search}

This section describes the design of the \textsc{Search} agent, which identifies the patients
most likely to be relevant to a clinical question. Because not every word in a question carries clinical
information, the agent first extracts its most clinically relevant signals as short, chart-compatible
phrases (Section~\ref{sec:search-planning}), then runs the search tool over the patient-level index and 
returns the top-ranked patients. The design rests on three choices: the retrieval unit, the index representation, and
the ranking algorithm, each compared against alternatives in Section~\ref{sec:ablations}.

\subsection{Query Planning}
\label{sec:search-planning}

Given a query $q$, an LLM produces a structured retrieval plan. It first
assigns the query one of three profiles (\texttt{cohort\_identity},
\texttt{focused\_evidence}, or \texttt{complex}), and this profile sets the
downstream retrieval budgets. It then emits two sets of chart-compatible terms:
core and related. Core terms are concepts explicitly required by the question. 
Related terms, at most five, are synonyms or standard clinical phrases that help
retrieve patients whose notes use different wording. Each term receives an importance
score from 1 to 10.
The planner removes interrogative wording and generic outcome language, 
but preserves distinctions
that affect clinical meaning, including anatomical site and meaningful
polarity (e.g., ``culture negative'' rather than ``culture'').
The plan is checked against a fixed schema before it is passed to the search tool (Appendix~\ref{app:prompts}).

\subsection{Patient-Level Index}
\label{sec:search-representation}

Longitudinal questions create a granularity mismatch for conventional note
retrieval. A diagnosis, treatment, and subsequent response may occur in
different encounters, so ranking notes independently can miss a patient whose
complete history is relevant but whose individual notes are only weak
matches. 
The \SearchAgent{} agent therefore makes the patient, rather than the note,
the retrieval unit. 

We consider three representations of each note: the raw note text,
a structured representation of clinical concepts extracted according to a fixed schema,
and a free-text summary. Raw notes contain templates, administrative content, and repeated history that dilute
lexical retrieval, while a fixed schema cannot capture the varied ways clinical concepts are expressed in real notes
or the complex relations between them. We therefore index summaries, produced offline by a lightweight language model
instructed to retain clinically meaningful events\footnote{For example, diagnoses, symptoms, medications, laboratory or imaging findings, and follow-up.} 
and changes while preserving uncertainty; the alternatives are compared in Section~\ref{sec:ablations}

For patient $p$, the indexed representation
concatenates the summaries of that patient's notes, preserving
encounter-level clinical events while allowing signals across time to
contribute to one score. Because summarization may introduce information loss or noise, 
they are used only for patient discovery; final answers are grounded in the original notes.

\subsection{\SearchAgent{} Tool}
\label{sec:search-algorithm}

The search tool supports three ranking algorithms: BM25 \citep{robertson2009bm25},
GREP-style literal matching, and a hybrid composition of the two. The final design uses
the hybrid algorithm; the alternatives are compared in Section~\ref{sec:ablations}.

The hybrid algorithm has two stages. In the first stage, BM25 searches the full
patient-summary index separately for each planned keyword or phrase, and each patient
is assigned the maximum score it achieves for any single phrase. The 1,000 highest-scoring
patients form the candidate set. In the second stage, a GREP-style step tests each planned
phrase as an exact contiguous string in the patient summaries (e.g., chronic kidney disease'' 
is matched as a phrase rather than as the individual words chronic'', kidney'', and disease''),
and re-ranks candidates by phrase coverage.

We score phrases independently, rather than concatenating them into a single query string, 
because questions typically focus on several parts of a patient's journey. A strong lexical match
on any single phrase can be a strong relevance signal, which a joint query
would dilute with phrases whose vocabulary the planner did not capture.
The second stage then promotes patients whose summaries contain exact mentions of a larger share of
the query's clinical concepts, while retaining BM25 as the scalable first-stage retriever.

The \SearchAgent{} agent returns the top-$k$ patients after re-ranking, where
$k$ is set by the assigned question profile
(Section~\ref{sec:search-planning}). 
These selected patients define the candidate set passed to \ResearchAgent{}, 
which uses their original notes to determine eligibility and generate the final answer.

\section{\ResearchAgent{} Agent}
\label{sec:research}

\ResearchAgent{} begins with three inputs: the user query, the patients
selected by \SearchAgent{}, and the weighted clinical keywords produced during
query planning. It looks through the notes of the selected patients; retrieval
summaries are used to find patients but are never treated as answer evidence.

The keyword weights determine which notes are read first. Let $s_k$ be the
planner-assigned importance of keyword $k$, let $N$ be the number of candidate
notes, and let $df_k$ be the number of candidate notes containing that keyword.
We adjust each keyword weight as
\[
w_k=s_k\max\left(0.25,
\frac{\log((N+1)/(df_k+1))}{\log(N+1)}\right).
\]
Thus, an important keyword receives a high score, while a keyword appearing in
many notes is downweighted because it is less useful for identifying the right
evidence. A note's score is the sum of $w_k$ over the distinct planned keywords
it contains. Repeating the same keyword does not increase the score. Notes that
match several important, relatively uncommon keywords therefore rank highest.
For each selected note, the system extracts a short window around the matched
text. An LLM reviews these windows and labels them fully relevant, partially
relevant, or irrelevant to the question. It discards irrelevant windows and
uses the retained evidence to determine which patients qualify and to write
the final answer. Every answer claim links back to the corresponding patient,
note, and source lines. If the available notes do not support an answer, the
system reports partial or insufficient evidence rather than filling the gap
from the retrieval summaries.

\section{Experimental Setup}
\label{sec:experimental-setup}

\paragraph{Data.}
Our primary development resource is a representative random sample drawn from
de-identified records covering approximately 140 million patients across the
United States. The sample contains approximately 2.1 million notes from
roughly 33{,}000 patients. To ensure that each record forms a meaningful
longitudinal journey, we include only patients with at least five clinical
notes. Journey length varies from 5 to 250 notes per patient. 
We also use an additional synthetic dataset of 8,343 notes 
from 321 patients (median 26 notes per patient), generated from clinical concepts 
extracted from de-identified source notes, for smaller-scale controlled experiments.
Appendix~\ref{app:reproducibility} provides more details.

\paragraph{Queries.}
Our queries are complex clinical questions designed so that no single note can
fully answer them: a satisfying answer requires combining evidence from
multiple encounters across a patient's journey. A subset is written by
clinical terminologists to reflect realistic investigative needs. The
remainder is generated bottom-up by sampling notes from one patient's journey
and prompting an LLM to compose a question whose answer requires evidence
across those notes. Queries are retained only when the corpus contains
adequate patient-level coverage.

\paragraph{\SearchAgent{} evaluation.}
The scale of the real-world corpus makes comprehensive gold labels infeasible
for every query. We therefore rely on an LLM judge (GPT-5.6-Sol~\cite{openai2026gpt56}) at
temperature 0. For \SearchAgent{}, each retrieved patient is evaluated independently by the LLM,
which receives the original query (not the planned keywords) with the patient's complete set of clinical notes.
A patient is considered only when the notes collectively provide sufficient evidence to satisfy all clinically
important requirements of the query.
The judge is blinded to the retrieval configuration, and we report judged patient precision at ranks 1, 5, and 10. 
Synthetic queries have comprehensive gold patient sets; we nonetheless apply the same judge
there for consistency. Due to the scale of the real-world corpus, 
judged scores cannot estimate recall against the
unknown complete cohort.

\paragraph{End-to-end evaluation.}
We evaluate the full \SearchAgent{}--\ResearchAgent{} pipeline on 20 
queries over the synthetic corpus. For each query, we compare the patients
cited in the final answer with the silver patient labels approved by an LLM,
and report patient precision, recall, and F1 score, macro-averaged over queries.
These single-run metrics measure whether the pipeline cites the right patients, not
whether every statement is clinically correct.

We additionally performed human evaluation using clinical experts to 
further assess the correctness of the system.
The human evaluation has two parts. Ten clinical experts first evaluate
patient and note selection for five synthetic cohort-identification questions;
each expert reviews two questions, and every patient and note receives four
reviews from four overlapping annotators. The experts then rate answers to 17
questions run over real clinical records for relevance and clinical accuracy.
Source records and evidence spans are withheld from this part for data-safety
reasons, and each answer receives two or three reviews. Both criteria use a
5-point scale with 5 as the best rating.
Appendix~\ref{app:human-evaluation-queries} provides the complete protocol,
synthetic questions, and question-level results.

\section{Results and Analysis}
\label{sec:results}

We evaluate \SearchAgent{} at scale on the proprietary corpus, report
end-to-end behavior on the synthetic corpus, and compare alternative designs
in Section~\ref{sec:ablations}.

\subsection{End-to-End Evaluation}
\label{sec:end-to-end-evaluation}

Table~\ref{tab:end-to-end-results} separates two evaluation settings. The
first uses LLM-approved silver labels for 20 synthetic questions, on which
the full pipeline achieves 0.87 patient precision, 1.00 recall, and 0.91 F1.
The second uses human evaluation. Clinical experts review evidence selection
for five synthetic cohort-identification questions, yielding 1.00 patient
precision and 0.97 note precision. Four reviews were submitted for each
patient and note, and all four selected the same binary label for every item;
we report this agreement descriptively. On 17 questions run over real clinical
data, all answer-relevance and clinical-accuracy ratings are 5/5.

\begin{table}[t]
\centering
\footnotesize
\setlength{\tabcolsep}{4pt}
\begin{tabularx}{\columnwidth}{@{}>{\raggedright\arraybackslash}X
  >{\raggedright\arraybackslash}p{2.15cm}
  >{\raggedleft\arraybackslash}p{2.35cm}@{}}
\toprule
\textbf{Evaluation} & \textbf{Measure} & \textbf{Result} \\
\midrule
\multicolumn{3}{@{}l}{\textit{LLM-approved silver labels (20 questions)}} \\
End-to-end synthetic performance & Patient P/R/F1 & 0.87/1.00/0.91 \\
\midrule
\multicolumn{3}{@{}l}{\textit{Human evaluation (5 questions)}} \\
Synthetic evidence selection & Patient precision & 1.00 (6/6) \\
& Note precision & 0.97 (35/36) \\
\midrule
\multicolumn{3}{@{}l}{\textit{Human evaluation (17 questions)}} \\
Real clinical data & Answer relevance & 5/5 across all annotator ratings \\
& Clinical accuracy & 5/5 across all annotator ratings \\
\bottomrule
\end{tabularx}
\caption{End-to-end synthetic and human evaluation. Real-data ratings use
a 5-point scale with 5 as the best score.}
\label{tab:end-to-end-results}
\end{table}

\subsection{\SearchAgent{} Performance}

On the proprietary corpus, patient-level summaries with hybrid ranking
obtain the highest reported means: 0.90 P@1, 0.90 P@5, and 0.93 P@10. These
preliminary aggregates do not form a fully paired comparison. For each
configuration and cutoff, the evaluator averages only queries that retrieved
at least $k$ unique patients; consequently, the effective query count can
differ across cutoffs and configurations. The historical aggregate retained
the means but not a common-query intersection. We therefore treat these values
as descriptive and do not attach uncertainty estimates or significance claims.

\section{Ablations}
\label{sec:ablations}

We ablate the \SearchAgent{} retrieval unit, indexed representation, and ranking algorithm.\label{sec:search-ablations} 
We vary three design choices of the \SearchAgent{} agent: the search algorithm, the index representation (raw notes, summaries, or structured concepts), 
and the index granularity (note-level versus patient-level). 
Configurations are compared by judged patient precision at ranks 1, 5, and 10 on the large proprietary corpus. 
Structred concepts are evaluated only on the synthetic dataset, because extracting them for the full proprietary corpus is costly.
Further details for each ablation are provided in Appendix~\ref{app:search-ablations}.

As shown in Table~\ref{tab:search-ablation}, hybrid ranking over patient-level summaries outperforms standalone 
BM25 and GREP using the same query keywords. Summaries also consistently outperform raw notes for both BM25 and GREP, 
and patient-level indexes formed by concatenating summaries generally outperform those formed from raw note content. 
These results suggest that summaries provide a stronger representation for lexical retrieval by preserving 
salient clinical signals while reducing noise from the full notes. 
Reported scores are averaged across three query files.

\begin{table}[t]
\centering
\small
\resizebox{\columnwidth}{!}{%
\begin{tabular}{lrrr}
\toprule
\multicolumn{4}{l}{\textit{\SearchAgent{} algorithm (patient-level summaries)}} \\
Hybrid & \textbf{0.900} & \textbf{0.900} & \textbf{0.930} \\
BM25 & 0.600 & 0.620 & 0.580 \\
GREP & 0.375 & 0.250 & 0.213 \\
\midrule
\multicolumn{4}{l}{\textit{Index representation (patient-level)}} \\
BM25, summary & 0.600 & 0.620 & 0.580 \\
BM25, raw notes & 0.400 & 0.220 & 0.130 \\
GREP, summary & 0.375 & 0.250 & 0.213 \\
GREP, raw notes & 0.333 & 0.178 & 0.111 \\
\midrule
\multicolumn{4}{l}{\textit{Index granularity}} \\
BM25, summary: patient & 0.600 & 0.620 & 0.580 \\
BM25, summary: note & 0.500 & 0.260 & 0.230 \\
BM25, raw notes: patient & 0.400 & 0.220 & 0.130 \\
BM25, raw notes: note & 0.500 & 0.260 & 0.240 \\
GREP, summary: patient & 0.375 & 0.250 & 0.213 \\
GREP, summary: note & 0.500 & 0.275 & 0.188 \\
GREP, raw notes: patient & 0.333 & 0.178 & 0.111 \\
GREP, raw notes: note & 0.375 & 0.175 & 0.150 \\
\bottomrule
\end{tabular}}
\caption{\SearchAgent{} ablations on the
proprietary clinical corpus.}
\label{tab:search-ablation}
\end{table}

\section{Conclusion} 
\label{sec:conclusion} 
Longitudinal clinical data contains comprehensive information about patient journeys, 
but reasoning over millions of notes is computationally challenging. 
Patient Insight addresses this by treating patients as the unit of retrieval 
and using compact note summaries as searchable representations. 
Its hybrid retrieval strategy combines fast BM25 search with more selective GREP-based matching, 
enabling patient discovery to scale while reserving expensive model context for 
evidence-grounded reasoning over relevant records. 

Our evaluation results demonstrate the feasibility of patient-level retrieval at 
scale while grounding final answers in the original clinical notes. In human
evaluation, the system achieves 1.00 patient precision and 0.97 note
precision. Answers produced for the real-data questions receive 5/5 for
relevance and clinical accuracy across all submitted ratings.
Together, these design choices provide a scalable and auditable framework 
for reasoning over longitudinal patient journeys.

\newpage
\section*{Limitations and Ethical Considerations}
\label{sec:limitations}

The reported experiments establish the feasibility of the approach rather than
providing clinical validation. They cover a limited set of datasets, workloads,
and questions. The large-scale \SearchAgent{} study uses
configuration-dependent query subsets and an LLM judge; the smaller human
study reports evidence verification for five synthetic cohort-identification
questions and answer ratings for 17 questions run over real clinical data. We
deliberately limit detailed evidence verification to five synthetic queries
because longitudinal cohort questions can span multiple patients and many
notes, making careful clinical review time-consuming. Even this focused set
required evaluation of 6 patient selections and 36 source notes, with four
overlapping expert reviews per item. We prioritized review quality over breadth
to assess usability with practicing clinicians and develop trust in
the framework's evidence selection and generated answers; these results should
not be interpreted as comprehensive clinical validation.

As with other systems that analyze clinical narratives, performance depends on
the quality and representativeness of the underlying data, the formulation of
the research question, and the available clinical documentation. The system is
query-driven rather than proactive: it generates insights in response to user
questions and does not independently identify potentially important patterns
or emerging clinical concerns. Future work should examine robustness,
fairness, calibration, and operational impact in prospective settings. Use of
clinical records must also remain subject to appropriate privacy protections,
access controls, institutional governance, and applicable regulatory
requirements. The proprietary corpus cannot be released and lacks exhaustive
cohort labels; its judged precision results therefore complement rather than
replace the reproducible controlled evaluations. The code and prompts will be
released upon acceptance, and the synthetic data will be released separately
with its companion dataset paper.

\bibliography{patient-insight-search}

\appendix

\newpage
\section{Prompt and Schema Details}
\label{app:prompts}

\paragraph{\SearchAgent{} planning contract.}
The \SearchAgent{} planner is instructed to preserve the user's question verbatim,
assign one of three query profiles, and generate chart-compatible retrieval
terms. Core terms must express concepts directly required by the question;
related terms are limited to five synonyms or standard clinical phrases. Each
term has an integer importance score in $[1,10]$. Question wording and generic
outcome language are excluded, while clinically meaningful polarity is
preserved. The planner returns:

\begingroup
\footnotesize
\begin{verbatim}
{
  "query": string,
  "query_type":
    "cohort_identity" |
    "focused_evidence" |
    "complex",
  "core_keywords": [
    {"keyword": string,
     "importance": 1..10}, ...],
  "related_keywords": [
    {"keyword": string,
     "importance": 1..10}, ...]
}
\end{verbatim}
\endgroup

Deterministic validation requires at least one core term, at most five related
terms, valid scores, and a supported query type. Case-insensitive duplicates
retain their maximum score.

\section{Reproducibility and Availability}
\label{app:reproducibility}

\paragraph{Data provenance.}
The proprietary corpus contains governed, de-identified longitudinal records
from seven health systems and cannot be released. Its approximately 33,000
patients and 2.1 million notes are used for evaluating \SearchAgent{} and \ResearchAgent{}.
Our synthetic corpus is generated from clinical concepts and their attributes,
extracted from de-identified source notes according to a predetermined schema, together with
the temporal relationships between them.
Source-note text is not supplied to the generator or reproduced in the released data.
Extracted representations are screened for residual identifying information,
and flagged representations are manually reviewed before generation.
The synthetic dataset contains 8,343 notes from 321 patients, with a median of 26 notes per patient.

\paragraph{Models and retrieval.}
The evaluated \SearchAgent{} and \ResearchAgent{} agents use GPT-5.6-Sol at
temperature 0. The \SearchAgent{} relevance judge uses GPT-5.4 at temperature
0, and offline note summaries use GPT-4.1-nano~\cite{openai2025gpt41} at temperature 0.
\SearchAgent{} uses BM25Plus with
$k_1=1.5$, $b=0.75$, and $\delta=1.0$. Tokenization lowercases text and uses
punctuation-aware splitting, retains stopwords, and discards tokens shorter
than two characters. Literal matching is case-insensitive substring matching
without stemming, morphological expansion, or built-in abbreviation
normalization. The frozen hybrid implementation retrieves 1,000 BM25
candidates using the maximum per-patient score across independently searched
planned phrases, ranks them by distinct planned-expression coverage, and
preserves BM25 order among ties. Indexes are rebuilt when source data change;
incremental updates were not evaluated.

\paragraph{Budgets and implementation.}
The planner assigns each question to a query category and applies corresponding
patient/note/window-character ceilings: 50/10/1,000 for cohort identity,
8/20/3,000 for focused evidence, and 15/20/3,000 for complex questions. The
window-character ceiling applies to each evidence window, not the complete
request. The end-to-end synthetic evaluation comprises 20 questions processed
by the full pipeline with these category-specific limits. The evaluated
prototype is preserved as a frozen experimental implementation; model
instructions are stored as declarative \SearchAgent{} and \ResearchAgent{}
agent files.

\section{Expanded Ablations and Examples}
\label{app:ablations}

\paragraph{Worked \SearchAgent{} plan.}
For the question \textit{``Find patients with ovarian cancer who are GIS
positive or HRD positive,''} the planner classifies the request as
\texttt{cohort\_identity} and produces:

\begingroup
\footnotesize
\begin{verbatim}
core:
  ovarian cancer (10)
  GIS positive (10)
  HRD positive (10)
related:
  genomic instability score positive (8)
  homologous recombination deficiency
    positive (8)
\end{verbatim}
\endgroup

The terms are retrieval hypotheses: the \textit{or} relation and patient
eligibility are checked against source-note evidence rather than inferred from
term occurrence alone.

\section{Human Evaluation Protocol and Queries}
\label{app:human-evaluation-queries}

\paragraph{Synthetic data.}
Ten clinical experts evaluate final system outputs. The evidence-selection
evaluation reported here contains five synthetic cohort-identification
questions. Each expert reviews two questions, and the balanced assignment
ensures that every displayed patient and note is reviewed by four experts.
Across the complete set, experts assess 6 patients and 36 associated notes.

\paragraph{Annotation interface and task.}
Cohort-identification questions require annotators to answer whether each patient is a 
correct member of the requested cohort and whether each displayed note is 
relevant to the question. 
All judgments are binary Yes/No choices, with no option preselected.
Each patient view contains the qualification
rationale, accepted note identifiers, and the final cited excerpts with note
identifiers and highlighted source-line numbers. Cohort-identification
annotators answer whether each patient is a correct member of the requested
cohort and whether each displayed note is relevant to the question.
All judgments are binary Yes/No choices, with no option preselected;
annotators may also provide optional comments.

\paragraph{Real data.}
The study also includes 17 questions run over real clinical records. For
data-safety reasons, this portion shows only the question and answer, without
patient records, citations, or evidence spans. Each real-data answer is
reviewed by two or three experts for relevance and clinical accuracy, yielding
40 ratings per criterion. Both criteria use a 5-point scale on which 5 is the
best rating and no response is preselected.

\paragraph{Aggregation and agreement.}
Human-verified precision divides the number of displayed patients or notes
labeled relevant by the total number displayed. The four submitted reviews for
each patient and note contain the same binary label, so majority and unanimous
aggregation produce identical results. We report this agreement descriptively.
For each real-data criterion, we report the number of reviews that selected the
highest score (5 of 5). Aggregate results appear in
Section~\ref{sec:end-to-end-evaluation}.

\begin{table*}[t]
\centering
\scriptsize
\begin{tabular}{@{}p{0.68\textwidth}rr@{}}
\toprule
\textbf{Question} & \textbf{Patient precision} &
\textbf{Note precision} \\
\midrule
Find newborns with birth-related cephalohematoma or scalp swelling who
subsequently developed clinically significant jaundice requiring NICU-level
monitoring or treatment. & 1/1 (100\%) & 9/10 (90.0\%) \\
Identify adults with asymptomatic monoclonal B-cell lymphocytosis
confirmed by flow cytometry after evaluation of persistent lymphocytosis or
abnormal white blood cell counts who are managed with hematology surveillance
rather than immediate treatment. & 1/1 (100\%) & 6/6 (100\%) \\
Identify adults with recurrent allergic upper-airway symptoms managed
in urgent care or primary care with antihistamines or intranasal steroids and
no evidence of acute infection. & 2/2 (100\%) & 7/7 (100\%) \\
Identify pediatric patients with community-acquired pneumonia who have
persistent focal infiltrates or ongoing pleuritic chest pain after initial
macrolide treatment and require repeat emergency evaluation with antibiotic
escalation. & 1/1 (100\%) & 2/2 (100\%) \\
Find patients with recurrent inflamed seborrheic keratoses of the scalp
or face managed with office-based cryotherapy and subsequent post-procedure
follow-up or repeat treatment. & 1/1 (100\%) & 11/11 (100\%) \\
\textbf{Overall (5 questions)} & \textbf{6/6 (100\%)} &
\textbf{35/36 (97.2\%)} \\
\bottomrule
\end{tabular}
\caption{Clinical-expert-verified patient and note precision for each
synthetic cohort-identification question.}
\label{tab:human-precision-by-case}
\end{table*}

\section{Search Ablations}
\label{app:search-ablations}
 
This appendix describes the alternatives compared in Section~\ref{sec:ablations}.
The \textsc{Search} design space has three decisions: the retrieval unit
(note or patient), the index representation (raw notes, summaries, or
structured clinical concepts), and the ranking algorithm (BM25, GREP, or
hybrid). In all configurations, the planning step and the planned terms are
identical; only the index and the ranking algorithm change.
Table~\ref{tab:design-space} summarizes the options, and
Table~\ref{tab:search-ablation} reports the results.

\begin{table}[h]
\centering
\small
\renewcommand{\arraystretch}{1.15}
\begin{tabularx}{\columnwidth}{@{}>{\raggedright\arraybackslash}p{2.35cm}
  >{\raggedright\arraybackslash}X@{}}
\toprule
\textbf{Decision} & \textbf{Options} \\
\midrule
Retrieval unit & Note index (one entry per note) \newline
  Patient index (one entry per patient, all notes joined) \\
\addlinespace[4pt]
Index representation & Raw note text \newline
  LLM summary per note \newline
  Structured concepts extracted with a fixed schema \\
\addlinespace[4pt]
Ranking algorithm & BM25 \newline
  GREP-style literal matching \newline
  Hybrid (GREP re-ranking of BM25 candidates) \\
\bottomrule
\end{tabularx}
\caption{\textsc{Search} design space. The final design uses a patient
index of summaries ranked with the hybrid algorithm.}
\label{tab:design-space}
\end{table}

\paragraph{Retrieval unit.}
In the note-level setting, every clinical note is indexed independently.
The planned terms are searched against individual notes, the
highest-scoring notes are retrieved, patient identifiers are recovered from
note metadata, and a selection function $F(\cdot)$ converts the top-ranked
notes into a ranked list of patients based on how many of their notes
appear in the top results.
This keeps notes as self-contained retrieval units and identifies exactly
which notes matched. However, a single note rarely covers a full
longitudinal question. A query that requires a diagnosis, a later
treatment, and a subsequent response may retrieve a highly relevant
diagnosis note or treatment note while missing the rest of the pattern, so
a relevant patient can rank low when each of their notes matches only one
part of the question.
 
In the patient-level setting, the representations of all of a patient's
notes are concatenated into a single index entry, and the search tool
ranks patients directly, with no $F(\cdot)$ step. Evidence from different
encounters contributes to the same score, which changes the retrieval
question from which note looks most similar to the query to which patient
history contains the clinical signals the query requires. This matches the
role of \textsc{Search}, which only needs to identify where
\textsc{Research} should look; \textsc{Research} then returns to the
patient's original notes to establish the answer.
 
\paragraph{Index representation.}
We compare three representations of each note. In every case, the
patient-level entry is the concatenation of the patient's note-level
representations.
 
\textit{Raw notes.} The source text is indexed directly. No information is
removed, but real-world notes contain substantial lexical noise: templated
sections, administrative content, history copied forward from earlier
encounters, and long lists of negative or normal findings (e.g., ``patient
denies chest pain,'' ``no shortness of breath''). This text adds
vocabulary unrelated to the query and inflates lexical scores for patients
with long records.
 
\textit{Summaries}. Each note is summarized offline by GPT-4.1-nano, which
is instructed to keep diagnoses, symptoms and their progression,
medications and treatment changes, laboratory and imaging findings,
procedures, plans, follow-up, and disposition, while dropping repeated
history, administrative detail, and lists of normal findings. The summary
is a retrieval-oriented representation rather than a reproduction of the
note. 
 
\textit{Structured concepts.} An LLM maps each note onto a predefined
schema of clinical concepts and their attributes, covering diagnoses,
medications, procedures, symptoms, laboratory and imaging findings,
devices, anatomy, and temporal information.
The extracted fields are linearized into text before indexing.
This makes clinical concepts explicit and consistent across notes written
in different styles. Its weakness is flexibility: the same concept is
expressed in many ways in real notes, and a fixed schema has limited
capacity to represent relations between concepts, such as which treatment
followed which diagnosis or what response it produced. The schema can
instead be discovered and refined from the corpus, where the LLM proposes new concepts and
attributes as it processes notes \citep{pham-etal-2024-topicgpt}. This adapts the schema to the dataset,
but the schema and its extraction prompt grow with every addition and
become specialized to the concepts seen during discovery. The structured
representation therefore trades coverage for explicitness, which matters
for longitudinal questions that depend on relations across encounters.
 
\paragraph{Ranking algorithm.}
All three algorithms receive the same planned core and related terms.
 
\textit{BM25} scores each indexed entry by planned-term frequency,
adjusted for how common each term is in the corpus and for entry length. 
As in the hybrid algorithm, each phrase is searched separately and an entry receives its
maximum score over phrases. BM25 is fast and produces a graded ranking,
but it scores individual tokens, so a multi-word concept such as ``chronic
kidney disease'' can be matched by unrelated occurrences of ``chronic''
or ``disease.''
 
\textit{GREP} tests each planned phrase as an exact, case-insensitive
substring of the indexed text and ranks entries by the number of distinct
planned phrases they contain.
Each entry, whether a patient or a note, is scored by query coverage: the share
of distinct planned phrases it contains, with higher-coverage entries ranked first.
Literal matching keeps multi-word clinical concepts intact and is precise
when the record uses the planner's wording, but it has no graded notion of
relevance, cannot match spelling or wording variants, and leaves many ties
on a corpus of tens of thousands of patients.
 
\textit{Hybrid} combines the two (Section~\ref{sec:search-algorithm}). BM25
first retrieves the 1,000 highest-scoring patients from the full index;
GREP then re-ranks this candidate set by distinct-phrase coverage,
preserving BM25 order among ties. BM25 supplies recall and a graded
ordering, and GREP promotes patients whose records contain exact mentions
of more of the question's concepts.
 
\paragraph{Evaluation protocol.}
Each configuration returns its top-ranked patients for the same questions,
and the LLM judge described in Section~\ref{sec:experimental-setup} labels each
returned patient as relevant or not, blinded to the configuration. We
report judged patient precision at ranks 1, 5, and 10, averaged over three
query files. A query contributes to P@$k$ only if the configuration returns
at least $k$ unique patients, so denominators can differ across rows and
cutoffs. For note-level indexes,
precision is computed over the patients produced by $F(\cdot)$, so the two
retrieval units are compared on the same patient-level output.

\subsection{Results on the Synthetic Dataset}
\label{app:search-synthetic}
 
Table~\ref{tab:search-ablation-synthetic} reports the same comparison on
the synthetic dataset, which also includes the structured-concept
representation. The representation finding matches the proprietary
corpus: at the patient level, summaries outperform raw notes and
structured concepts for both BM25 and GREP. Structured concepts rank
between the two, even though the synthetic notes are generated from
extracted clinical concepts, a setting that favors the structured index.
The granularity finding differs: on the synthetic dataset, note-level
retrieval matches or exceeds patient-level retrieval, most clearly for raw
notes. A possible reason is that synthetic notes are cleaner and more
self-contained than real notes, so a single note more often covers the
question; on the noisier proprietary corpus, patient-level indexing
performs better (Table~\ref{tab:search-ablation}). We do not evaluate
the hybrid algorithm on the synthetic dataset: its first stage keeps the top 1,000
BM25 candidates, which on a corpus of this size would retain most patients and reduce the hybrid to 
GREP re-ranking of nearly the entire corpus.
 
\begin{table}[h]
\centering
\small
\begin{tabular}{lllrr}
\toprule
\textbf{Unit} & \textbf{Algorithm} & \textbf{Representation} & \textbf{P@1} & \textbf{P@5} \\
\midrule
Patient & BM25 & Summary    & \textbf{0.53} & \textbf{0.78} \\
Patient & BM25 & Raw notes  & 0.45 & 0.63 \\
Patient & BM25 & Structured & 0.47 & 0.72 \\
\cmidrule(lr){1-5}
Patient & GREP & Summary    & \textbf{0.50} & \textbf{0.74} \\
Patient & GREP & Raw notes  & 0.39 & 0.61 \\
Patient & GREP & Structured & 0.44 & 0.69 \\
\midrule
Note & BM25 & Summary    & 0.59 & 0.77 \\
Note & BM25 & Raw notes  & \textbf{0.61} & \textbf{0.78} \\
Note & BM25 & Structured & 0.52 & 0.73 \\
\cmidrule(lr){1-5}
Note & GREP & Summary    & \textbf{0.53} & \textbf{0.80} \\
Note & GREP & Raw notes  & 0.41 & 0.65 \\
Note & GREP & Structured & 0.46 & 0.72 \\
\bottomrule
\end{tabular}
\caption{Judged patient precision for \SearchAgent{} ablations on the
synthetic dataset, by retrieval unit, ranking algorithm, and index
representation. Bold marks the best representation within each unit and
algorithm. Hybrid configurations have not yet been evaluated on this set.}
\label{tab:search-ablation-synthetic}
\end{table}

\clearpage

\end{document}